# Beyond Imitation: A Life-long Policy Learning Framework for Path Tracking Control of Autonomous Driving

Cheng Gong, *Graduate Student Member, IEEE*, Chao Lu, *Member, IEEE*, Zirui Li, *Graduate Student Member, IEEE*, Zhe Liu, Jianwei Gong, *Member, IEEE,* Xuemei Chen

*Abstract*—Model-free learning-based control methods have recently shown significant advantages over traditional control methods in avoiding complex vehicle characteristic estimation and parameter tuning. As a primary policy learning method, imitation learning (IL) is capable of learning control policies directly from expert demonstrations. However, the performance of IL policies is highly dependent on the data sufficiency and quality of the demonstrations. To alleviate the above problems of IL-based policies, a lifelong policy learning (LLPL) framework is proposed in this paper, which extends the IL scheme with lifelong learning (LLL). First, a novel IL-based model-free control policy learning method for path tracking is introduced. Even with imperfect demonstration, the optimal control policy can be learned directly from historical driving data. Second, by using the LLL method, the pre-trained IL policy can be safely updated and fine-tuned with incremental execution knowledge. Third, a knowledge evaluation method for policy learning is introduced to avoid learning redundant or inferior knowledge, thus ensuring the performance improvement of online policy learning. Experiments are conducted using a high-fidelity vehicle dynamic model in various scenarios to evaluate the performance of the proposed method. The results show that the proposed LLPL framework can continuously improve the policy performance with collected incremental driving data, and achieves the best accuracy and control smoothness compared to other baseline methods after evolving on a 7 km curved road. Through learning and evaluation with noisy real-life data collected in an off-road environment, the proposed LLPL framework also demonstrates its applicability in learning and evolving in real-life scenarios.

*Index Terms*—Autonomous driving, life-long learning, learning from demonstration, model-free control, path tracking,

## I. Introduction

ACCURATE path tracking control is crucial for autonomous vehicles to drive safely in dynamic and complex environments [1-4]. Early path tracking control methods based on static linear models or expert knowledge, such as PID control, pure-pursuit control [5], and etc. [6, 7], can work well when driving at low speed. For driving with higher speed, model predictive control (MPC) [1, 8], preview control [1], and other optimal control methods [9] have gained more attention and application for their accurate modeling of vehicle dynamics. However, the model accuracy is still limited, where small-angle assumption and model linearization are performed to easy the computational burden. Such inaccuracy restricted intelligent vehicles from driving in dynamic and complex environments. To improve model accuracy and adapt to environmental changes, much research has focused on extracting and estimating the nonlinear and time-relevant model parameters using posterior or online knowledge. In [10, 11], fuzzy-based methods are employed to estimate the model uncertainty and improve the robustness. A recurrent neural network is applied in [12] to predict vehicle motion. And Gaussian process is also employed in [13] for system dynamics learning. By cooperating with expert knowledge and online observation, these methods can more accurately capture the model dynamics and thus achieve better control performance. However, these models require dedicated modelling for the driving environment, where the estimation may fail when the uncertain parameter exceeds or faces multiple uncertain parameter presence.

Instead of estimating model parameters, many researchers also try to improve model adaptability by learning parametric models directly through imitation learning (IL) [14-17] or reinforcement learning (RL) [18-20]. Such parametric models can be highly adaptive to different environments, but they are naturally associated with high training costs and may introduce human error or driver difference [21] into the collected data. To eliminate human error from policy learning, a conditional imitation scheme is proposed in [22]. Instead of learning from human demonstrations, an end-to-end policy for high-speed off-road driving is proposed in [14], which is trained by imitating a state-of-the-art MPC controller. In [23], the authors used an actor-critic RL method to learn parameters for adaptively coordinating a PP and PID controller. In [19], a lane following method based on DDPG with double critic networks is proposed, which learns to drive on a circular road at high speed. DDPG is also combined with kernel learning for vehicle path tracking in [20] and achieves high accuracy and smoothness. Although these parametric policies are very adaptive to scenarios with rich training data, they degrade when faced with new scenarios, where their high learning cost also aggravates the burden of generalization to new scenarios.

This work was supported in part by the National Science and Technology Major Project (2022ZD0115503), in part by National Natural Science Foundation of China under Grant 52372405, in part by the BIT Research and Innovation Prompting Project (Grant No.2023YCXY004), and in part by the Key Lab of Intelligent Unmanned Systems, Beijing Institute of Technology. (*Corresponding author: Chao Lu, Jianwei Gong*).

The authors are with the School of Mechanical Engineering, Beijing Institute of Technology, Beijing 100081, China. (e-mail: chenggong@bit.edu.cn; chaolu@bit.edu.cn; z.li@bit.edu.cn; 1120180980@bit.edu.cn; gongjianwei@bit.edu.cn; chenxue781@bit.edu.cn).

Zirui Li is also with the Chair of Traffic Process Automation, "Friedrich List" Faculty of Transport and Traffic Sciences, TU Dresden, 01069 Dresden, Germany.

Apart from enlarging the dataset for better scene coverage, enabling intelligent agents to learn and evolve with accumulative experience is shown to be a more appealing solution for policy generalization and improvement. RL-based methods naturally support continual learning in execution where their knowledge can be acquired by exploration and interaction with reward feedbacks. However, they require enormous trials and great learning costs, and the verge of a sub-optimal policy through non-heuristic exploration can be extremely inefficient. Combining learning from demonstration with online fine-tuning for policy can on a level alleviate such problems as in [24, 25]. But random exploration and insufficient learning may still degrade the policy performance in the online learning process. A hypernetwork [26] and a dynamic confidence value [27] are proposed to alleviate the performance degradation that occurs in the online learning process of RL. However, they are task-specific and require explicit task discrimination and corresponding task labels, which can be difficult to acquire for continuous control tasks.

Instead of learning through exploration and interaction, directly learning from incremental data can be more efficient and applicable for online tasks. This learning scheme is called life-long learning (LLL) or incremental learning [28, 29], where the agent can generalize to new tasks with incremental execution knowledge. Many solutions have been proposed to achieve the LLL scheme, where one main focus is to prevent the forgetting problem caused by network overfitting to incremental data [30]. The idea of LLL makes it very applicable to intelligent robots, where robots are expected to continually explore new environments or tasks and continually learn to adapt to new ones [31]. Several recent applications are presented in navigation [32], behavior prediction [33, 34], and place recognition [35]. These researches provide great insights into the practical benefits of employing LLL for continuous tasks, where the model can be improved continuously in execution. However, these proposed learning schemes need to learn from ground truth data in new tasks, which can be difficult to acquire in continuous control tasks online. In general, traditional LLL is not capable of learning policy with no ground truth data, while RL suffers from the inefficient learning process.

To tackle the aforementioned problems and enable learning-based path-tracking policy to evolve and fine-tune its performance with accumulated driving experience, this paper proposes a life-long policy learning (LLPL) framework that enables efficient and continuous policy learning and guaranteed performance improvement in online execution. To achieve such goals, an efficient imitation learning scheme is introduced for initializing policy for LLPL. The proposed scheme enables great adaptability with minimum tuning or data collection requirements, which is realized by learning the policy with mappings from historical state transition to action. To safely improve the policy with a small amount of incremental execution knowledge, an efficient LLL method that does not necessarily require task labels in the learning process is utilized. Thus, the LLPL can achieve continual policy learning for path tracking control with incremental data, which enables the policy to adapt and evolve with online execution knowledge. By cooperating with a knowledge evaluation scheme for the incremental knowledge in LLPL, redundant and inferior knowledge can be filtered to avoid performance degradation in the continuous learning process, hence guaranteeing policy performance improvement in the learning with incremental data. The main contributions of this paper are as follows:

- An efficient model-based policy learning method for path tracking is proposed. No explicit vehicle parameter nor perfect demonstration is needed for policy training and a sub-optimal policy can be learned with only minutes of driving data, where the policy is formulated as an inverse optimal control law that can be learned with only historical IMU and steering data.
- The life-long policy learning (LLPL) framework for vehicle control is proposed, which enables a pre-trained policy to evolve safely and fine-tune in continuous driving tasks. Thus, reducing the data required for policy initialization and adapting the policy with realistic data feedback.
- A knowledge evaluation and updating method for optimized policy learning is proposed, enabling the policy to evolve and adapt without performance degradation and avoid control fluctuation. By employing the knowledge evaluation method, both knowledge distribution and optimality in incremental knowledge and memory are managed and optimized for safe continual learning.

The LLPL framework architecture is shown in Fig.1, which contains two major parts, separately policy imitation learning and policy life-long learning. An initial policy is first learned from demonstration data. Then the initialized policy will be deployed in the execution environment and continuously improved with evaluated incremental knowledge. The paper is organized as follows. In section II, the policy initialization part is introduced. In section III, the policy life-long learning with knowledge evaluation part is introduced. Section IV presents the experiments and evaluation of LLPL. Finally, the conclusion and future directions are given in Section V.

## II. LEARNING POLICY FROM DEMONSTRATION

Generally, the vehicle lateral dynamics with vehicle state $x$ and steering angle $\delta$ can be formulated as:
$$\dot{x} = f(x, \delta). \tag{1}$$

In vehicle lateral control problems, the primary task of the control policy is to operate the vehicle to smoothly track a collision-free reference trajectory provided by the upper planner. Thus, given a reference trajectory $X_{\text{ref}}$, the cost function over an infinite time horizon is usually designed as:

$$J = \int_0^\infty \left( \|x(t) - X_{\text{ref}}(t)\|_Q + \|\delta(t)\|_R \right) dt,$$
$$x(t) = x_0 + \int_0^t f(x, \delta) dt \tag{2}$$

where $\|\cdot\|_Q$ and $\|\cdot\|_R$ are separately the L-2 norm with $Q$ and $R$ as the weight identity matrixes to balance the accuracy and the smoothness of the policy. The vehicle lateral control problem can then be formulated into an optimal control problem. Model predictive control methods are most commonly used in practice to solve this problem numerically. However, the MPC methods are computationally costly, while their performances are highly reliant on the determination and estimation of algorithm and vehicle parameters. To ease the computational burden and free policy from parameter tuning, a policy learning method for learning a direct mapping from vehicle state to optimal control value will be proposed. By shrinking the optimal control problem to a time-discrete problem with a fixed time window $\Delta t$, the corresponding receding horizon optimal control policy can be acquired:

$$\delta^* = \arg\min_{\delta} \left( \left\| x + \int_0^{\Delta t} f(x,\delta)dt - X_{\text{ref}}(\Delta t) \right\|_Q + \|\delta\|_R \right). \quad (3)$$

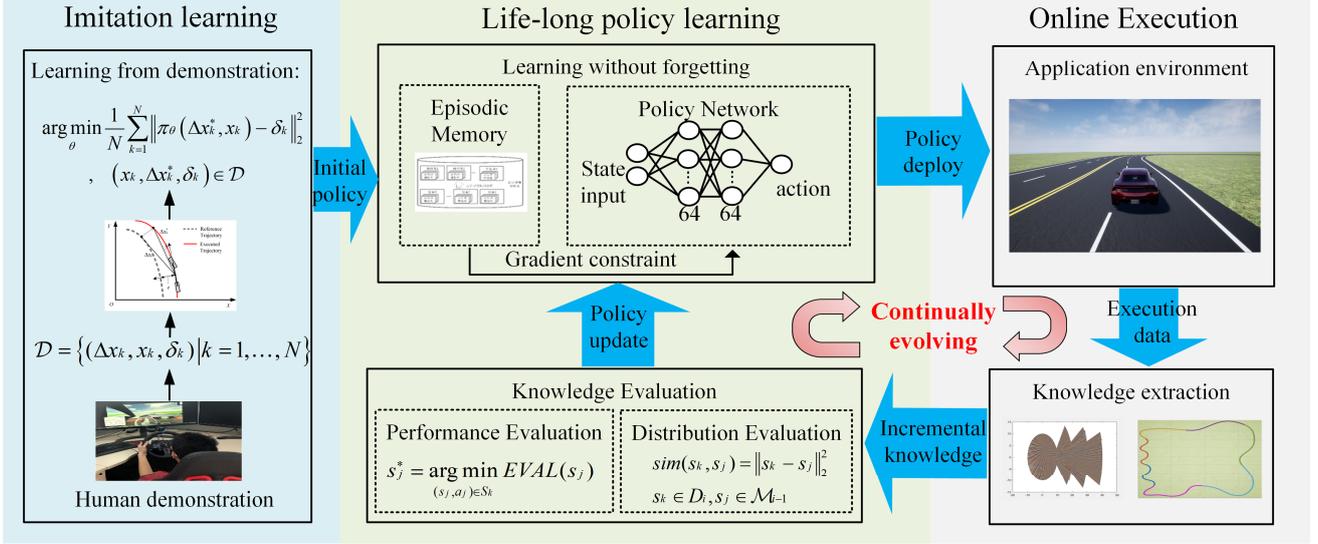

**Fig.1.** The illustration of the LLPL framework. First, demonstration data will be used for policy initialization through imitation learning. Through analyzing historical state transition and corresponding control, even unperfect demonstration can be used for policy initializing. Then, the initialized policy is deployed to the executing environment where the policy can be updated and evolved online through learning incremental driving knowledge collected in execution. By employing a knowledge evaluation scheme and maintaining an episodic memory of previously learned knowledge, the policy is updated with evaluated data and a gradient constraint that ensures performance improvement over updating. Thus, enabling the policy to adapt to the execution environment and continually fine-tune its performance with online execution knowledge.

To approximate the policy from demonstration, the extracted vehicle lateral control knowledge can be described with a dataset containing historical vehicle state and corresponding control action with sampling time $T$:

$$\mathcal{D} = \{(\Delta x_k, x_k, \delta_k) | k = 1, \ldots, N\}, \quad (4)$$

where $\Delta x_k = X_{\text{ref},k} - x_k$ is the control target regarding reference trajectory $X_{\text{ref},k}$ and vehicle state $x_k$ in step $k$, $\delta_k$ is the corresponding steering action, and $N$ represents the total amount of extracted samples from demonstrations. However, as shown in Fig.2, the actual control $\delta_k \propto (\Delta x_k, x_k)$ doesn't always correspond to the optimal control $\delta_k^* \propto (\Delta x_k, x_k)$ due to the imperfect demonstration. To avoid learning these errors, instead of imitating the control policy directly from the demonstration of tracking the reference trajectory, we can learn the policy directly from historical vehicle responses. Assume the lateral control to be constant in the fixed time window $\Delta t$, by integrating (1) we have:

$$x_{k+\Delta t/T} - x_k = \int_0^{\Delta t} f(x_k, \delta_k)dt, \quad (5)$$

By inverse (5), we can obtain:

$$\delta_k = f^{-1}(\Delta x_k^*, x_k), \quad (6)$$

where $f^{-1}$ is the inverse dynamics function and $\Delta x_k^* = x_{k+\Delta t/T} - x_k$ represents the state transition in the fixed time window $\Delta t$. By replacing the vehicle's historical control target $\Delta x_k$ with the vehicle's historical state difference $\Delta x_k^*$, the cost of control $\delta_k$ in (3) can be described as:

$$J(x_k, x_k, \delta_k) = \|0\|_Q + \|\delta_k\|_R. \quad (7)$$

Hence the optimal control $\delta_k^* \propto (\Delta x_k^*, x_k)$ can be approximated by $\delta_k$ when control effort is not considered with $R = \mathbf{0}$. And vehicle historical operational trajectory can be regarded as ideal path tracking knowledge without tracking error. Thus, the optimal control policy can be approximated by learning the inverse dynamic function $f^{-1}$:

$$\pi_\theta = \arg\min_\theta \frac{1}{N} \sum_{k=1}^{N} \left\| \pi_\theta(\Delta x_k^*, x_k) - \delta_k \right\|_2^2, \quad (x_k, \Delta x_k^*, \delta_k) \in \mathcal{D}. (8)$$

By retracing the historical vehicle state space and action space, the proposed policy learning method is independent of corresponding task knowledge from demonstration and is capable of learning path tracking knowledge with imperfect demonstration. However, it is worth noting that the control smoothness cannot be guaranteed by the demonstration as indicated in (7), where control error is excluded but control effort is not minimized. To tackle this problem, the distribution of the demonstration can be modified to adjust the driving style if enough demonstrations are acquired.

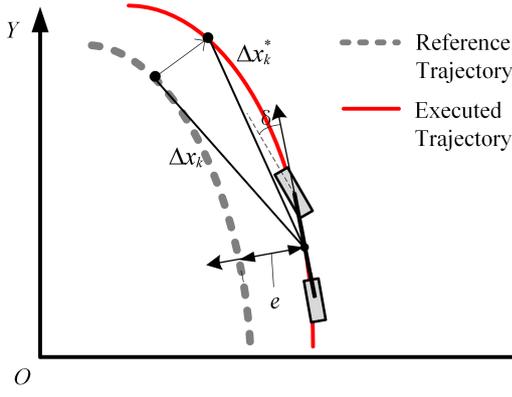

**Fig. 2.** Vehicle path tracking control with amended tracking target.

## III. LIFE-LONG POLICY LEARNING WITH INCREMENTAL KNOWLEDGE

To reduce the learning cost and fully exploit the online execution knowledge, we introduce the LLL scheme to supplement the IL-based policy learning scheme in policy refining and updating. Ideally, similar to human drivers that can learn to drive better with increasing driving experience, the autonomous driving policy is expected to evolve and improve with execution knowledge acquired online and eventually verged on optimal performance over all scenarios. However, policy updating with incremental knowledge usually requires the policy to be retrained on overall data, including already learned knowledge and cumulative knowledge, which can be extremely time-consuming and inefficient. The inconvenience in previous policy updating and retraining is mainly due to the catastrophic forgetting problem that may happen in the process of learning incremental knowledge, where fitting the new distribution of incremental knowledge may result in a mismatch with previously learned knowledge, leading to the forgetting of learned knowledge.

The LLPL framework is proposed in this section to ease the catastrophic forgetting problem and enable the policy to evolve with increasing driving experience efficiently. Specifically, a gradient descent constraint method is applied in the learning procedure to prevent the policy from forgetting already learned knowledge while learning new knowledge. A knowledge evaluation scheme is proposed to assess and optimize the incremental knowledge in order to guarantee performance improvement over policy iteration and reduce the learning cost.

### A. Average Gradient Episodic Memory

To avoid forgetting knowledge from previous tasks when learning incremental knowledge, a life-long learning algorithm, Average Gradient Episodic Memory (A-GEM) [36], is employed in our framework. It is an improved version of Gradient Episodic Memory (GEM) [37] and is more computationally efficient than its predecessor, which makes it more applicable to on-platform learning that demands fast learning and a low computational cost.

To avoid catastrophic forgetting, A-GEM maintains an episodic memory to store knowledge sampled from previous tasks. The episodic memory will be used to estimate the loss of previous knowledge when learning new knowledge. It is worth noting that incremental knowledge is separated by different labels into different tasks as A-GEM is initially oriented only for classification tasks, which is different from policy learning with continuous knowledge. In this study, the tasks are defined as a series of time-continuous knowledge collected under a specific time duration. And to avoid forgetting, the algorithm prevents the loss of previous knowledge from increasing when decreasing the training loss on the new tasks. Instead of computing the episodic memory loss of every previous task in the training process, A-GEM computes the average episodic memory loss to approximate the loss on all previous tasks, and the object of A-GEM for learning a new task $t$ can be expressed as:

$$\min_{\theta} l(f_\theta, \mathcal{D}_t) \quad s.t. \quad l(f_\theta, \mathcal{D}_t) \leq l(f_\theta^{t-1}, \mathcal{M}), \quad (9)$$

where $f_\theta$ is the model with parameter $\theta$, $\mathcal{D}_t$ is the data from the new task $t$, and $f_\theta^{t-1}$ is the model trained till task $t-1$, $\mathcal{M} = \cup_{k<t} \mathcal{M}_k$ is a randomly sampled batch of the episodic memory of previous tasks in which $\mathcal{M}_k$ is the episodic memory of the $k_{th}$ learned task. The optimization problem of the loss in (9) can be reduced to the optimization of the model gradients, where the constraint on loss can be transferred to the constraint on gradient descent direction. The expected gradient $\tilde{g}$ for decreasing training loss should be in the same direction as the gradient for reducing episodic memory loss:

$$\min_{\tilde{g}} \|\tilde{g} - g\|_2^2 \quad s.t. \quad \tilde{g}^\mathsf{T} g_{\text{ref}} \geq 0, \quad (10)$$

where $g = \partial l(f_\theta, \mathcal{D}_t)/\partial \theta$ is the gradient calculated in training the current task, and $g_{\text{ref}} = \partial l(f_\theta, \mathcal{M})/\partial \theta$ is the reference gradient calculated using the randomly sampled batch of the episodic memory $\mathcal{M}$. Compared to GEM, which solves the optimization problem in (10) through the quadratic program (QP), A-GEM adopts a more effective solution, which derives the solution when directions of two gradients contradict via:

$$\tilde{g} = g - \frac{g^\mathsf{T} g_{\text{ref}}}{g_{\text{ref}}^\mathsf{T} g_{\text{ref}}} g_{\text{ref}}, \quad (11)$$

where the gradient can be directly computed and is very time-efficient compared to solving a QP problem.

### B. Life-Long Policy Learning with Knowledge Evaluation

For the policy $\pi_\theta^{i-1}$ pre-trained with demonstration data $\mathcal{D}_d$, directly updating the policy with newly collected driving data $\mathcal{D}_i$ may lead to forgetting of previously learning knowledge, which leads to direct performance degradation. To alleviate such problem, the A-GEM is integrated in our framework for safe policy incremental learning. To ensure the stability in updating policy with new driving data, an episodic memory $\mathcal{M}_{i-1}$ will be first initialized with randomly sampled data from $\mathcal{D}_d$. The episodic memory can be seen as a sparse representation of previously learned knowledge. And to fully exploit the usage of episodic memory, we introduce a knowledge evaluation scheme to first evaluate the increment knowledge $\mathcal{D}_i$ before policy updating to avoid repetitive learning and distribution conflict. To measure the distribution, Euclidian distance will be calculated between each sequence of data from episodic memory $\mathcal{M}_{i-1}$ and incremental $\mathcal{D}_i$:

$$sim(s_k, s_j) = \|s_k - s_j\|_2^2, \quad s_k \in \mathcal{D}_i, s_j \in \mathcal{M}_{i-1} \quad (12)$$

where $s_k = (x_k, \Delta x_k^*)$ represents the input state of policy. If data in incremental data $\mathcal{D}_i$ is similar to data in episodic memory $\mathcal{M}_{i-1}$, then the quality of knowledge will be compared. The quality of knowledge refers to the control performance of corresponding historical data, where in this paper the performance is evaluated by control effort. As shown in (7), the initially learned policy is only approximately optimal for accurate trajectory tracking, which failed to minimize the control effort. To improve the policy performance and reduce control fluctuation, the data in $\mathcal{D}_i$ with greater control effort than previously learned knowledge will be omitted. It is worth noting that other indicators, such as lateral acceleration, can also be chosen for different considerations like riding comfort. Thus, the data with different distributions or have lesser control effort are selected from the incremental data $\mathcal{D}_i$:

$$\mathcal{D}_i^* = \{\{s_k, \delta_k\} | sim(s_k, s_j) \geq \eta_D, \forall s_j \in \mathcal{M}_{i-1}\} \\ \cup \{\{s_k, \delta_k\} | sim(s_k, s_j) \leq \eta_D, \delta_k^2 \leq \delta_j^2, \forall s_j \in \mathcal{M}_{i-1}\}, \quad (13)$$

where $\mathcal{D}_i^*$ is the evaluated and screened incremental data, and $\eta_D$ is the similarity threshold to evaluate and screen the incremental data. By employing the A-GEM and $\mathcal{D}_i^*$, the previous policy $\pi_\theta^{i-1}$ can be updated safely with more samples from unseen state space or better performance:

$$\pi_\theta^i = \arg\min_\theta l(\pi_\theta^{i-1}, \mathcal{D}_i^*) \\ s.t. \quad l(\pi_\theta^{i-1}, \mathcal{D}_i^*) \leq l(\pi_\theta^i, \mathcal{M}_{i-1}), \quad (14)$$

where the goal is to update policy $\pi_\theta$ with incremental data $\mathcal{D}_i^*$ while avoiding instability and forgetting in learning, and $l(\cdot)$ is the mean square error loss in (9) as:

$$l(\pi_\theta^{i-1}, \mathcal{D}_i^*) = \frac{1}{N}\sum_{k=1}^N \|\pi_\theta^{i-1}(s_k) - \delta_k\|_2^2, \quad (s_k, \delta_k) \in \mathcal{D}_i^*. \quad (15)$$

As shown in (14), the episodic memory from previous policy updating will be applied to constrain the learning direction in the following policy updating epoch. Thus, properly updating the episodic memory is vital to ensure the consistency of policy updates. Firstly, learned knowledge should be evenly distributed in episodic memory to ensure the fidelity of the episodic memory. Similar to the knowledge evaluation of incremental data $\mathcal{D}_i$, Euclidian distance will be calculated between each sequence of data from episodic memory $\mathcal{M}_{i-1}$ and $\mathcal{D}_i^*$:

$$sim(s_k, s_j) = \|s_k - s_j\|_2^2, \quad s_k \in \mathcal{D}_i^*, s_j \in \mathcal{M}_{i-1} \quad (16)$$

Data are only sampled to episodic memory if they share no similar distribution with episodic memory $\mathcal{M}_{i-1}$ or have better performance, so as to improve the storage efficiency and optimality of the episodic memory. By ensuring the optimality of episodic memory, the policy performance can be guaranteed to improve over iterations. The knowledge evaluation will be conducted when knowledge from incremental knowledge $\mathcal{D}_i$ contradicts prior knowledge stored in the episodic memory $\mathcal{M}_{i-1}$, and the knowledge with a better evaluation score will be selected as:

$$s_j^* = \arg\min_{(s_j,\delta_j)\in S_k} EVAL(s_j), \quad (17)$$

where $EVAL(\cdot)$ is the evaluation function, and $S_k$ is the set that stores knowledge that contradicts in distribution:

$$S_j = \{\{s_j, \delta_j\} | sim(s_k, s_j) \leq \eta_M, \exists s_j \in \mathcal{M}_{i-1}, \forall s_k \in \mathcal{D}_i^*\}, (18)$$

where $\eta_M$ is the similarity threshold to ensure the knowledge is evenly distributed in episodic memory, and can be adjusted according to the storage demand for episodic memory, the evaluation function is similar to the evaluation of incremental data:

$$EVAL(s_k) = \|\delta_k\|_2^2. \quad (19)$$

Thus, by supplementing the knowledge that is not represented in $\mathcal{M}_{i-1}$, and the knowledge that has better performance, we can obtain new episodic memory after each batch of incremental knowledge:

$$\mathcal{M}_i = \{s_j^*\} \cup \mathcal{M}_{i-1}, \quad s_j^* \in S_k. \quad (20)$$

With the constraint from the episodic memory in both knowledge distribution and knowledge quality, the proposed life-long policy learning method can evolve with accumulated incremental execution knowledge in both generalization ability and task performance. The whole procedure of LLPL is described in Algorithm 1.

| **Algorithm 1**: LLPL framework |
|---|
| **Input**: Demonstration $\mathcal{D}_d$, $\pi_\theta(s_k)$ with random initialized $\theta$, $\mathcal{M} \leftarrow \varnothing$ |
| // policy initialization through IL |
| 1: $\pi_\theta = \arg\min l(\pi_\theta, \mathcal{D}_d)$ |
| 2: initialize $\mathcal{M}$ with random samples from $\mathcal{D}_d$ |
| 3: $t = 0, \mathcal{D}_i \leftarrow \varnothing$ |
| 4: **while** policy execution **do**: |
| 5:     sample $s$ from environment |
| 6:     execute steering $\delta = \pi_\theta(s)$ |
| 7:     collect incremental data $\mathcal{D}_i \leftarrow \mathcal{D}_i \cup \{s, \delta\}$ |
| // policy online updating through LLPL |
| 8:     **if** $t \geq t_{update}$ **do**: |
| 9:        evaluate incremental data with $\mathcal{M}$ |
|         $\mathcal{D}_i^* = \{\{s_k, \delta_k\}|sim(s_k,s_j) \geq \eta_D, \forall s_j \in \mathcal{M}\}$ |
|         $\cup \{\{s_k,\delta_k\}|sim(s_k,s_j)\leq\eta_D, \delta_k^2 \leq \delta_j^2, \forall s_j \in \mathcal{M}\}$ |
| 10:       update the policy with life-long learning |
|         $\pi_\theta^* = \arg\min_\theta l(\pi_\theta, \mathcal{D}_i^*) \quad s.t. \quad l(\pi_\theta, \mathcal{D}_i^*) \leq l(\pi_\theta^*, \mathcal{M})$ |
| 11:       $\pi_\theta \leftarrow \pi_\theta^*$ |
| 12:       updated $\mathcal{M}$ with equation (14)-(18) |
| 13:       $t = 0$, $\mathcal{D}_i \leftarrow \varnothing$ // clear for next update |
| 14:     **end if** |
| 15: **end while** |
| **Output**: $\pi_\theta$, $\mathcal{M}$ |

## IV. EXPERIMENT AND EVALUATION

To evaluate the proposed LLPL framework, this section presents the procedure and results in three groups of experiments for validating policy learning performance. Firstly, the data collection procedure and the constructed experimental environment are described. Secondly, a group of experiments in two typical driving scenarios is presented. The policies are evaluated by revisiting the same scenario, and the results will be discussed from three major aspects concerning IL performance, LLL performance, and the aid of knowledge evaluation. Thirdly, the policy learning framework and the learned policy will be examined and evaluated in a more complex scenario where their performance will also be compared with an online reinforcement learning method and MPC. Lastly, we examine the applicability of LLPL by training the policy with noisy real-data, where the performance and continuous learning ability of LLPL is evaluated and discussed.

### A. Experimental Environment and Data Collection

Due to safety concerns for the potential policy failure, the evaluation and comparison experiments will only be conducted in the simulated environment, and the experiment settings are shown in Fig.3. MATLAB/ SIMULINK is used for algorithm deployment and simulation. A human driver is asked to drive the vehicle in the simulated environment through a Logitech G29 steering input to acquire human demonstration for initial policy learning with IL. Enlightened by how human drivers learn to drive, the task for the driver is to cruise with statistical speeds of 5m/s, 10m/s, 15m/s, and 20m/s in a specific area, where he should operate as varied as possible to acquire as much control knowledge as possible. Since the proposed policy learning scheme does not necessarily require expert path tracking demonstration, the only requirement for the driver is to perform different control actions with varying vehicle states that can lead to more prosperous vehicle control knowledge distribution. The computing device used for experiments is equipped with Core i7-10875H and RTX2060 with 16G RAM. A neural network with two hidden layers and 64 units per layer is applied to approximate the policy $\pi_\theta$.

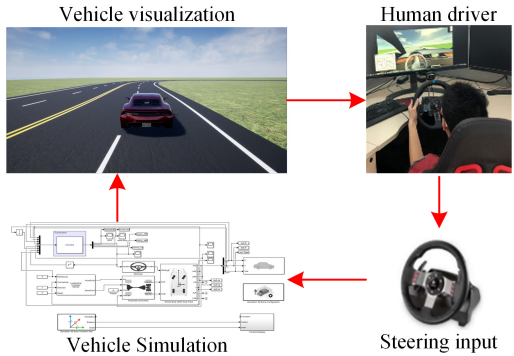

**Fig.3.** The simulation environment and human demonstration data collection procedure.

### B. Evaluation of Policy Learning Performance

To evaluate the learning performance of LLPL, a double lane-change driving scenarios shown in Fig.4 (a) is employed. As previously mentioned, 10 minutes of human driving demonstration are processed and applied for initial policy training. Firstly, to evaluate the effectiveness of the proposed IL scheme, the initial policy will be deployed to the testing scenario with a cruise speed of 12m/s, where the driving data at this speed is not included in demonstrations. As for evaluating the continual learning performance of the proposed LLL scheme, the initial policy is updated through the LLPL framework and deployed to revisit the same scenario. After the execution of the initial policy, the data of the first travel will be used to fine-tune the initial policy through LLPL. The fine-tuned policies are then deployed to revisit the test scenario. The results are shown in Fig.4 (b)-(c), where the fine-tuned policy after execution of the initial policy is marked as 1st Revisit, and the fine-tuned policy after execution of 1st Revisit as 2nd Revisit.

As the results show, the initial policy can perform well in the testing scenario. Despite the lack of demonstration in the testing cruising speed, the IL policy is capable of fitting the path tracking knowledge with few imperfect demonstrative data. This shows the capability of IL in efficient policy learning through direct mapping vehicle dynamic responses with reduced vehicle state space. However, few demonstrative data may lead to sub-optimal policy approximation, for which the LLPL framework is proposed further to improve the policy performance with incremental operation knowledge. As the results illustrate, the performance of the 1st and 2nd Revisit outrank the initial policy in both lateral and heading accuracy, which proves that the LLPL scheme can improve the policy performance with acquired incremental execution data after execution. As shown in Fig.4 (c), the lateral deviation of 1st Revisit is reduced compared to the initial policy. It can be further reduced in the 2nd Revisit, indicating that the policy can moderate itself through LLPL to better match the actual vehicle responses. The average deviation of the 2nd Revisit is reduced by 23.78% compared to the 1st policy and 66.76% compared to the initial policy. This ability of continual learning will enable the autonomous vehicle to learn and evolve on the go. The policy will only need a relatively small amount of data to initialize and can keep refining itself with actual execution data.

As shown in equation (5) and (6), the IL method learns an inverse dynamics function, where the convergence of $x_k \rightarrow x_{k+\Delta t/T}$ can be guaranteed with $\delta_k *$ when the inverse dynamics function is properly trained for corresponding state and state transition. To evaluate the actual convergence performance, the policies trained in double lane change are evaluated, and results are presented in Fig.5. shows the stability and convergence performance of the IL and LLPL policy in continuous learning. The IL and LLPL can both converge smoothly and with incremental knowledge, LLPL can also present smoother performance than the initial IL policy.

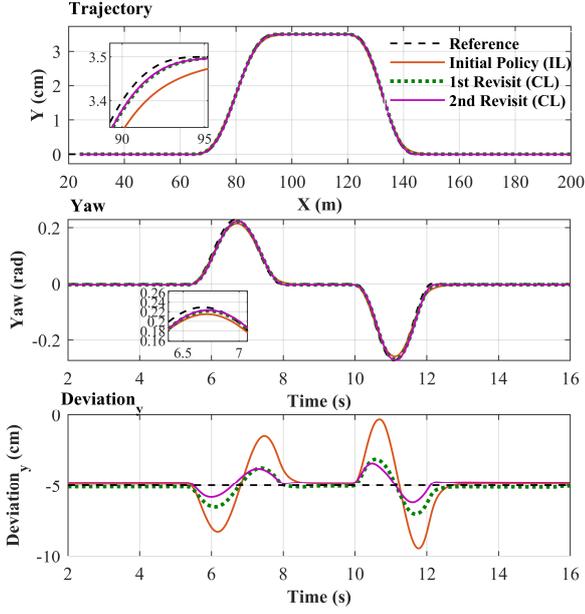

**Fig.4.** The experimental results in the double lane change.

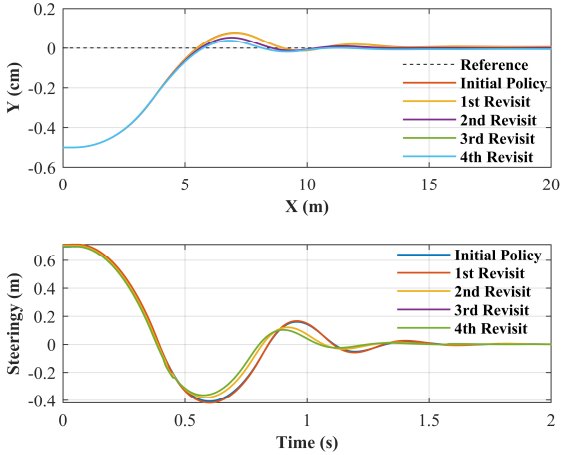

**Fig.5.** The experimental results in the double lane change.

To further analyze the continual learning performance of LLPL in policy updating, offline IL and traditional LLL norms are deployed and compared in the same double lane-changing scenario iteratively for six epochs. For the IL method, iteration means retraining the policy after each execution with initial training data and all collected execution data to avoid catastrophic forgetting. For LLL and LLPL, the policy will be updated with an LLL scheme that updates the policy with only newly acquired incremental execution data. Different from LLPL, LLL employs only A-GEM, where knowledge is not evaluated and episodic memory is randomly sampled from learned data. The specific settings of three compared different learning methods are:

1) **Offline imitation learning (IL)**: policy trained offline with historical driving data and collected new data. After each execution, the accumulated execution knowledge is supplemented with the collected human driving data, and the extended data are employed for policy retraining with the IL scheme to prevent forgetting.

2) **Direct life-long learning (LLL)**: Direct LLL scheme using A-GEM, where the task is defined as each execution. After initialized with IL, the policy is directly upgraded through A-GEM with newly collected driving data. No knowledge evaluation is applied and the memory is randomly sampled from learned data, which means the policy is updated with all knowledge collected in the execution and try not forgetting even the worst of the knowledge. The sample ratio is set to 10% (randomly sample 10% data from execution data).

3) **Proposed LLPL (LLPL)**: The proposed LLPL learning framework with knowledge evaluation and memory updating. IL will be used for initial policy training with collected human driving data. After each execution, new data acquired from execution is first evaluated and screened before updating the policy through the learning scheme shown in Algorithm 1.

Results of the three learning methods are presented in Fig.6, where we directly show the root mean squared error (RMSE) in each iteration epoch of the trajectory deviation for better visualization. All three methods achieved similar performance in the first execution with the same initial training data. Although all compared methods can prevent forgetting, both IL and LL methods cannot steadily benefit from incremental execution knowledge. Compared to the IL and LL methods, the LLPL method shows better steadiness in performance improvement with incremental knowledge. This difference between LLPL and traditional LLL methods is caused by the difference in their learning scheme. Aim to not forget, IL employs all driving data that include historical data and incremental data, where conflicted data and data with varied performance will undermine the policy performance. As for traditional LLL, it shows worse policy performance in the incremental learning process compared to IL, which traded the forgetting ratio for lower computational cost. Compared to the traditional "not to forget everything" LLL paradigm, the proposed LLPL is capable of evaluating incremental knowledge in order to avoid conflicted knowledge and evaluate incremental knowledge regarding performance, guaranteeing performance improvement in the online incremental learning process.

Regarding data storage requirements, the size increment of data and memory used in each epoch are illustrated in Fig.7. For LLPL, the size increment of data storage (episodic memory size) is the execution data after knowledge evaluation in each epoch. As for LLL, the size increment is the fixed amount of randomly selected operation data in each epoch. As Fig.7 shows, the increment of knowledge for LLPL decreases after each epoch, while the other two compared methods do not. This result explains the ability of knowledge evaluation to adjust the distribution of incremental knowledge, where the learned knowledge will be excluded to avoid over-fitting and minimize the storage requirement for data. As for learning efficiency, the policy initial training time of LLPL and LLL (about 400s) takes two times longer than the IL method (about 200s) since the initial episodic memory sampling with knowledge evaluation is conducted. However, since only execution data will be used, LLPL and LLL only need less than 4s to update the policy with about 16s of incremental data, which is 100 times faster than the IL-learning method. Furthermore, the policy updating time of LLPL decreases after each epoch, where the learning cost is lessened by knowledge evaluation in more mastered scenarios. The significantly reduced learning cost in operation also indicates the potential of our proposed LLPL in the application for online policy learning and updating.

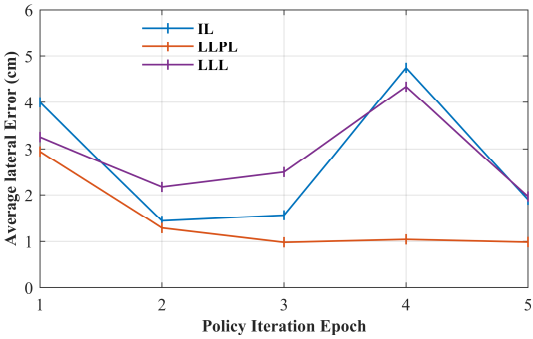

**Fig.6.** Average lateral deviation of different policy iteration epochs in the double lane-change scenario.

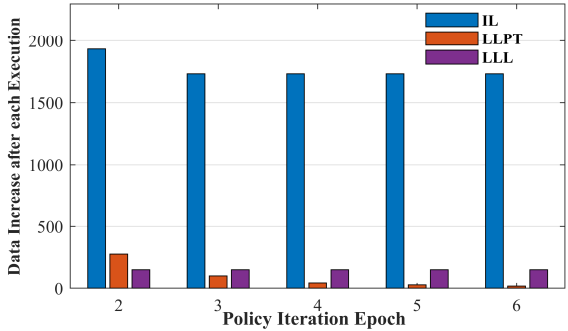

**Fig.7.** Data storage increment after execution of different policy iteration epochs in the double lane-change scenario.

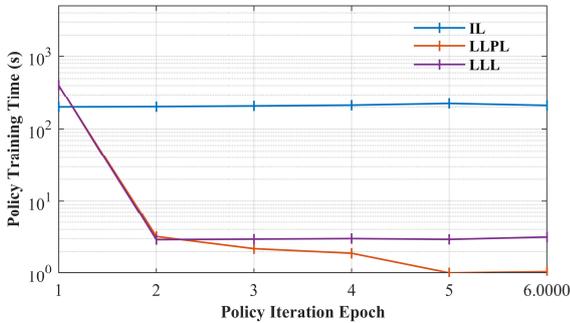

**Fig.8.** Policy training time cost of different policy iteration epochs in the double lane-change scenario.

## C. Policy Generalization in More Complex Scenario

In the previous part, we demonstrated that the policy could better its performance by revisiting identical scenarios and learning from incremental execution knowledge. However, the actual operating environment can be more complex and wildly varied. The policy will need to adjust to different scenarios instead of revisiting the same scenarios repeatedly. Thus, to evaluate the generalization ability of the proposed LLPL framework to learn and evolve in more complex and continuous scenarios, a curved road environment is adopted for further policy deployment with rich road characteristics and a total length of over 7km. The reference trajectory is extracted based on interpolating a set of selected waypoints manually selected from the center of the inner lane, as shown in Fig.9. To apply the LLPL, and an initial policy is trained the same way as previously mentioned, which is learned through the proposed IL scheme with 10 minutes of human driving demonstration.

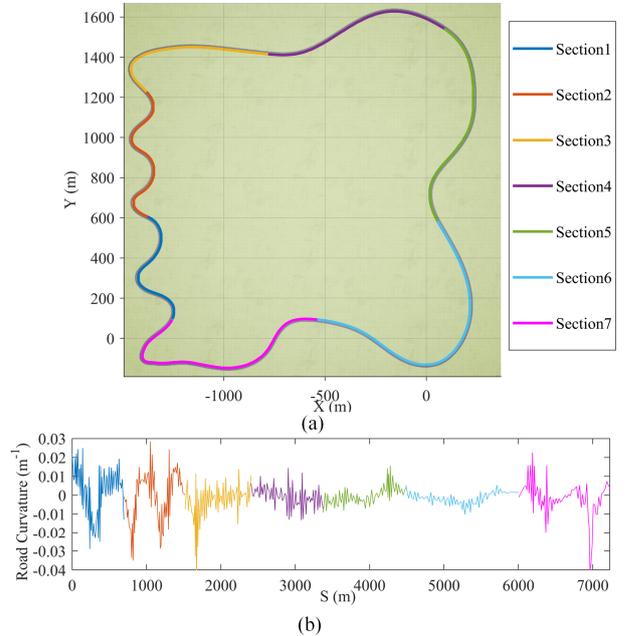

**Fig.9.** The curved road environment with seven segmented road sections, (a)the bird view of the whole trajectory, (b) the road curvature of different segmented sections.

The experimental environment and trajectory with segmentation are shown in Fig.9, where seven segmentations are separated and marked from Section 1 to Section 7 (for simplification, we use S1 – S7 to represent each section in the rest of the paper). Each section is approximately 1 km long, and the policy will be updated after each section with the execution knowledge obtained from the traveled section. For instance, the initial policy will be applied in S1, where the accumulated execution knowledge will be evaluated and used for updating the policy. The updated policy will then be applied in S2, and the execution knowledge from S2 will be used to update the policy for S3. Similarly, in other segments, the policy will be updated with accumulated execution knowledge for execution in the following segments.

As different sections of the curved road scenario vary greatly, the policy execution difficulty also differs. Thus, The IL-only method is also compared in this experiment to give a more straightforward reference for the proposed LLPL framework. In this experiment, the IL-only method shares the same initial policy as LLPL for better comparison in policy continual learning ability. A similar online RL framework used in [25] is also compared and analyzed in this experiment, which used offline demonstrations for policy initialization and fine-tuning the policy with online reinforcement learning. Similar to LLPL, the RL policy is also initialized with the same 10 minutes of human driving demonstration, and is updated after the execution of each section.

The loss of the critic network is calculated as:

$$L_C(\phi, \mathcal{D}_i) = \frac{1}{N}\sum_{k=1}^{N}\left(y_k - Q_\phi(s_k, \delta_k)\right), \quad (s_k, \delta_k) \in \mathcal{D}_d \bigcup \mathcal{D}_i, \quad (21)$$

where $y_k = r_k + \gamma Q_{\phi'}(s_{k+1}, \pi_{\varpi'}(s_{k+1}, \delta_{k+1}))$, $Q_\phi$ is the critic network parameterized by $\phi$ and $\pi_\varpi$ is the actor network parameterized by $\varpi$, $r_k = -e_{y,k}^2 - e_{yaw,k}^2 - \delta_k^2$ is the reward for each step, $\gamma$ is the discount factor, $\phi'$ and $\varpi'$ are the parameter of the target network. The loss of actor is calculated as:

$$L_A(\varpi, \mathcal{D}_i) = L_{BC}(\varpi, \mathcal{D}_i) + \lambda L_{PG}(\varpi, \mathcal{D}_i), \quad (22)$$

where $L_{BC}(\varpi, \mathcal{D}_i)$ and $L_{PG}(\varpi, \mathcal{D}_i)$ are separately the behavior cloning loss and policy gradient loss:

$$L_{BC}(\varpi, \mathcal{D}_i) = \sum_{k=1}^{N}\left\|\pi_\varpi(s_k) - \delta_k\right\|_2, (s_k, \delta_k) \in \mathcal{D}_d \bigcup \mathcal{D}_i, \quad (23)$$

$$L_{PG}(\varpi, \mathcal{D}_i) = -\sum_{k=1}^{N}\left\|Q_\phi(s_k, \pi_\varpi(s_k))\right\|_2, (s_k, \delta_k) \in \mathcal{D}_d \bigcup \mathcal{D}_i \quad (24)$$

and $\lambda$ is the hypermeter to ensure the learning stability in fine-tuning. The actor network is of the same structure as the LLPL, and critic network is composed of four layers of fully-connected neural network with 128 units of cells each. Worth noting that, we do not replicate the Q-filter used in the original work since we find it may lead to policy failure in a later section. To avoid the control fluctuation in online policy finetuning, the exploration noise is set to be 10% of the maximum steering value. And the policy updating in the first two sections employs the warm-up setting to ensure stability in finetuning, where $\lambda = 0$ and only behavior cloning loss will be used to update the actor. We empirically chose $\lambda = 0.05$ for policy learning since larger $\lambda$ usually leads to policy failure when updating in later sections.

Besides, a commonly used MPC method with a linearized vehicle dynamic model is also selected as the baseline method, which numerically solves the optimal control problem in (2) with a receding horizon scheme:

$$\min J(k) = \frac{1}{2}\sum_{i=k}^{k+N_p}\xi^T(i)\mathcal{Q}\xi(i) + \mathcal{R}\delta^2(i)$$

$$s.t. \quad \xi(i+1) = \mathcal{A}\xi(i) + \mathcal{B}\delta(i) \quad (25)$$

$$\delta_{\min} \leq \delta \leq \delta_{\max}$$

where $\xi = [y - y_{\text{ref}} \quad \dot{y} - \dot{y}_{\text{ref}} \quad \psi - \psi_{\text{ref}} \quad \dot{\psi} - \dot{\psi}_{\text{ref}}]^T$, $N_P$ is the predictive horizon, $\mathcal{A} \in \mathbb{R}^{4\times4}$ and $\mathcal{B} \in \mathbb{R}^4$ are the coefficient matrices determined with vehicle parameters, $\mathcal{Q} \in \mathbb{R}^{4\times4}$ and $\mathcal{R} \in \mathbb{R}^4$ are the positive definite weight matrices for balancing tracking accuracy and smoothness. The vehicle parameters used in coefficient matrices $\mathcal{A}$ and $\mathcal{B}$ are set as identical to the simulated vehicle platform, and the preview horizon $N_P$ is set to 50 steps with a sampling period of 0.1s.

The results of policy deployment in the curved road environment are presented in Fig.10, where the continuous results are separated with corresponding sections for better interpretation. All compared methods achieve similar performance in S1, which again proves the effectiveness of the proposed IL scheme in learning the vehicle control policy with only minutes of human demonstration. As S1 and S2 are both series of sharped curves with similar driving difficulty, the continual learning ability of LLPL can be straightforwardly evaluated by comparing with the IL-only method in S1 and S2. With incremental execution knowledge and knowledge evaluation, the lateral tracking deviation of LLPL policy is significantly reduced compared to the IL-only method in S2. The execution knowledge in S2 can further improve the policy performance through LLPL in S3, which can be observed more explicitly in Fig.11. Compared to the IL-only policy, lateral tracking error can be reduced by 85.6% and 75.7% in S2 and S3 by applying the proposed LLPL scheme. Different from the LLPL, the performance of RL degrades after fine-tuning with incremental knowledge. The performance degradation of RL is likely caused by the inaccurate value estimation since the states for traveling in 12m/s are not shown in the demonstration and the amount of incremental driving data is insufficient for accurate value estimation. This indicates that LLPL is more efficient in learning and fine-tuning policy with small amount of incremental data.

To further validate the adaptability of the proposed LLPL framework, the cruise speed is changed to 20m/s in more smoothed sections, including S4-S6. Although the IL-only policy reveals acceptable performance in middle-speed cruising on S1-S3, the policy begins to fluctuate when cruising at high speed on S4. As shown in Fig.10, the steering control of the IL-only policy trembles on S4-S6, leading to significant deviation in the vehicle heading and large lateral acceleration. And as for RL, the policy can maintain a stable steering at high speed but holds a steady state error after speed increases. Although cruising at 20m/s has been shown in the demonstration, the historical knowledge may be forgotten in the fine-tuning process of RL. The RL lateral error increases in S5 after fine-tuning with incremental data collected in S4, which may due to the inaccurate estimation of the value function and knowledge forgetting. Through knowledge evaluation of both data distribution and optimization, the proposed LLPL policy can avoid fluctuation and forgetting and provide a more stable performance than other methods. As Fig.10 shows, the LLPL policy can control the vehicle with similar or even better performance as the MPC, and the control smoothness of the LLPL policy is greatly improved in S5-S6 compared to S4 as a result of policy fine-tuning.

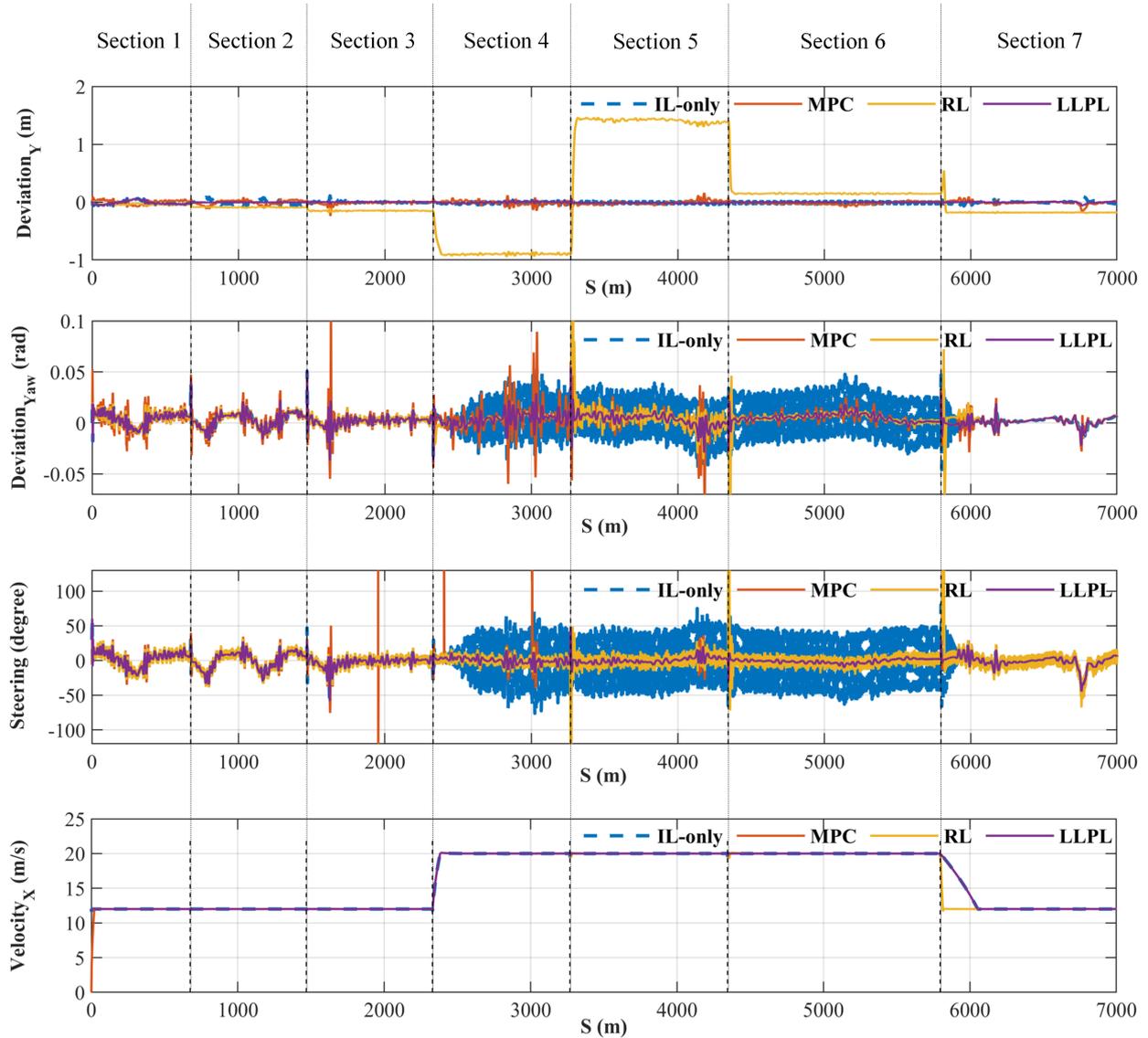

**Fig.10.** Experimental results of policy evaluation in curved road, where the results within corresponding sections are separated with dash lines for better visualization.

Regarding the data efficiency and effectiveness in the continuous road environment, more illustrative results are shown in Fig.11. Compared to IL-only, the RL policy after fine-tuning with incremental knowledge continues to degrade in S1-S5, and only sees some improvement in S5-S7, which is still worse than the initial policy. The cause of such performance degradation can be twofold. On the one hand, the initial policy performs well enough that any random exploration can undermine its performance. On the other hand, the low data efficiency severely limits the ability of the RL to improve in online fine-tuning. The large state-action space limits its ability to accurately approximate the value function for continuous control with only a small amount of demonstration and incremental data. Compared to the RL and IL methods, LLPL can safely improve the policy performance with a small amount of driving data and avoid fluctuation or forgetting. Compared to the initial policy, LLPL can reduce the overall average lateral and heading tracking error in the curved road by 46.82% and 41.22%, respectively. The robust online incremental learning capability can efficiently evolve a suboptimal policy in execution, reducing the data requirement for initial policy training and improving the overall safety and performance in the incremental learning process. The high data efficiency can also enable the policy to adapt and generalize online to environments with greater uncertainty or perturbations, such as slippery roads with varying adhesion coefficients. However, such an application is beyond the scope of this paper, where the focus is mainly on introducing the LLPL framework for general environments.

As shown in the above results, it is clear that LLPL has a better performance in continuous learning and adaptation to new road environments than RL. The superior performance of LLPL is achieved due to its better data efficiency, which benefits from the lifelong learning graph and the knowledge evaluation scheme. To further discuss the data efficiency of LLPL compared to RL or LLL without knowledge evaluation scheme, a fixed scene is selected to perform continuous learning for these policies, where they are updated and evaluated with data collected in each iteration. Thus, the data efficiency can be evaluated by improving the performance of the iterations required in the continuous learning. Section 4, shown in Figure 9, is used as the fixed scenario, and the corresponding experimental setup is identical to the previous experiment. All policies have the same initial policy, and the RMSE of the displacement in each iteration is shown in Fig.12.

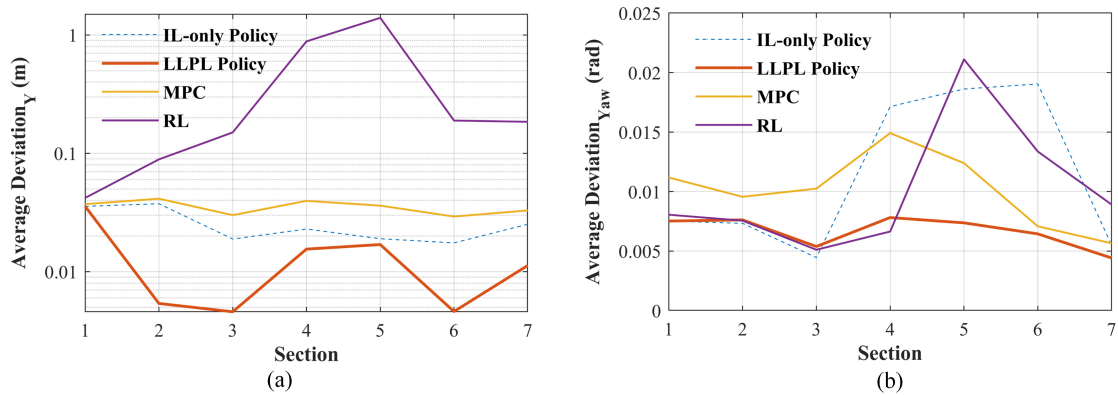

**Fig.11.** The average deviation of compared methods in different road sections, (a) lateral deviation, (b)heading deviation.

As the result shows, the LLPL can converge to a very low RMSE already in the second iterations, which means that it is able to adapt to the scenario with data from only one iteration. On the contrary, the RL degrades in the first few iterations, which is identical to the results shown in Fig.10 and Fig.11. Also, due to random exploration and overfitting, the performance of RL is more erratic than that of LLPL, although a demonstrative dataset is used for RL. In terms of data efficiency, it is clearly shown that RL can only evolve to a similar performance as LLPL after almost 100 epochs, using almost 100 times more data than LLPL. This further demonstrates that LLPL is more applicable compared to RL in policy continuous learning, where LLPL adapts the policy to its optimal performance in new tasks or environments faster than RL.

The results also show that for continuous control tasks, the LLL learning scheme deteriorates in the first few iterations because the quality of the new knowledge is not guaranteed and the data distribution difference can be large. And LLL converges to a similar performance as LLPL with almost ten epochs, which is similar to Fig. 6. Thus, the performance of LLL only depends on the overall quality of all the knowledge learned for the policy, and the policy performance can also degrade when learning in a new scenario. This demonstrates the necessity and benefit of introducing the knowledge evaluation scheme in LLPL, which significantly improves the data efficiency and online learning capability for improving policy performance. Nevertheless, LLL is still much more data efficient than RL in continuous learning, proving that the IL-based policies are potentially more efficient in online policy learning.

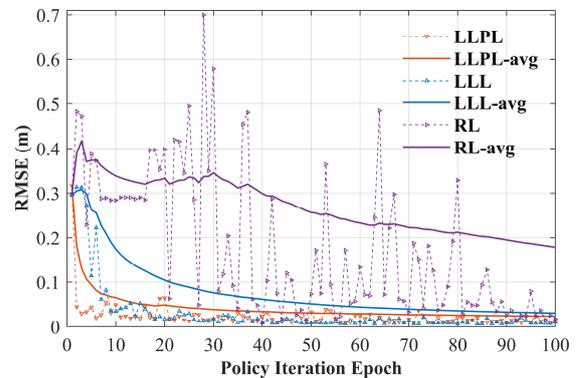

**Fig.12.** The continual learning performance of policies in a fixed scenario. The dashed line indicates the RMSE of the current iteration and the solid line indicates the average RMSE from the beginning to the current iteration.

*D. Learning with Real-Life Data*

Different from simulation, data collected by real-life vehicles tend to be noisy and inaccurate, which may lead to inaccurate approximation in the policy online learning procedure. The difference may also prevent the policy from learning and evolving when deployed in realistic driving environments, even leading to critical policy failure. To address such concerns, this experiment is designed to evaluate the applicability of the proposed LLPL framework in learning and evolving with collected real-life data. To cover the most extreme scenarios, the employed data in this experiment is collected on off-road rough terrain with a Jeep Wrangler. Only IMU and steering data are needed for training and updating the policy. The sampling frequency of IMU and steering data are separately 100Hz and 10Hz. A partial observation of the sampled data is presented in Fig. 13, where the noise and disturbance of the real-life data are presented.

Policies are trained with real-life data and evaluated in simulation. One section of the collected traversed trajectory is extracted to replicate the scenario in simulation. The parameters of the simulated vehicle are set to mimic the real vehicle, for instance, the steering ratio, wheelbase, mass and et al. But certain characteristics such as tire model and mass center are estimated and may not be accurate. Two policies are trained and evaluated, which separately are trained with IL and LLPL framework. Both policies are initialized with 36 minutes of real-life driving data. As for the LLL procedure in LLPL, the policy is updated each 100s with real-life historical data from corresponding traversed trajectories to reproduce the deployment in real-life. The two policies are executed in the simulated environment, and the results are shown in Fig.14 (b)-(c). Despite the noises in data and the dynamic differences between real vehicles and simulation platforms, both policies are capable of executing the path tracking task without failure. Although replayed real-life data are not equivalent to the true vehicle response in simulated evaluation environment, LLPL is still capable of exploiting the incremental data and improving the policy performance. The overall deviation of LLPL is reduced by 23.30% and 21.64% for lateral deviation and heading deviation compared to IL. Although the control fluctuation occurs due to data noises and mismatches between real-life data and simulation, it is significantly alleviated by LLPL through knowledge evaluation.

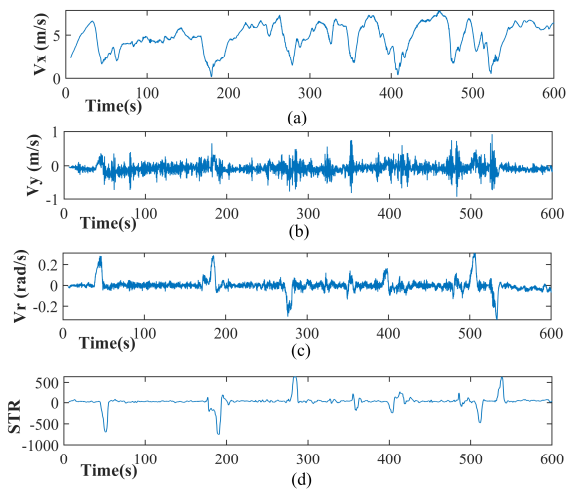

**Fig.13.** A segment of collected real life data on off-road terrain.

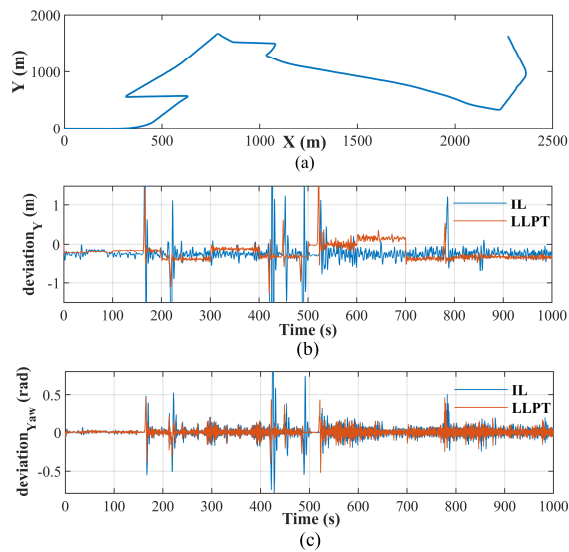

**Fig.14.** The performance of policy trained with real-life data in simulation: (a) the segment of the extracted path for policy evaluation, (b) the lateral deviation of evaluated policies, (c) the heading deviation of evaluated policies.

## V. CONCLUSION AND FUTURE DIRECTIONS

In this paper, the LLPL framework is proposed for autonomous vehicle path tracking control, which can efficiently learn an initial control policy from only a few minutes of demonstrations, and can also continuously update the policy with small amounts of incremental execution knowledge. By adopting the proposed knowledge evaluation method, the proposed LLPL can avoid learning redundant knowledge and improve control smoothness compared to IL. Experiments are conducted using a high-fidelity vehicle dynamic model to evaluate the continuous learning performance. Experimental results show that the proposed LLPL can efficiently learn an initial policy by IL and continuously improve it with a small amount of incremental knowledge. The LLPL policy is also evaluated on a curved road with a total length of over 7 km, where the LLPL is also compared with three baseline methods IL, RL and MPC. The results show that LLPL is able to safely learn and fine-tune with incremental driving data, while the performance of IL and RL may degrade. By using the knowledge evaluation scheme, the LLPL policy also achieves the best accuracy and control smoothness compared to other baseline methods. Real-world data collected in off-road terrain is also used to train the policy and evaluate it in simulation. Despite a slight performance degradation, the LLPL policy is able to learn and improve with incremental real-world data, demonstrating its applicability to real-world learning.

The main goal of the proposed LLPL framework is to enable the autonomous driving system to efficiently learn a model-free driving policy and to continuously adapt and evolve during execution, thus enabling rapid policy deployment for different vehicles and environments and significantly reducing the policy learning cost. However, there are still some limitations of the current work, which would provide some interesting directions for future work. First, the implicit dynamic model introduced in this work may not be sufficient to represent more complex driving scenarios, such as off-road environments, where pitch angle and road characteristics should also be considered. Second, although the performance improvement is guaranteed by the knowledge evaluation in LLPL, the performance of the initial policy may fail in out-of-distribution cases. Ensuring a lower performance bound for the initial policy can greatly reduce the data requirements of the initial policy and increase the generalizability of the LLPL framework. Finally, the proposed LLPL framework can also be generalized to other continuous control or even decision-related tasks, where how to formulate an efficient policy model and corresponding knowledge evaluation scheme remains an interesting future direction.

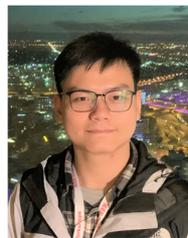

**Cheng Gong** (Graduate Student Member, IEEE) received the B.S. degree in mechanical engineering from Beijing Institute of Technology, China, in 2020. He is currently pursuing the Ph.D. degree in Beijing Institute of Technology, China. His research interests include intelligent vehicles, motion planning and control, machine learning, and life-long learning.

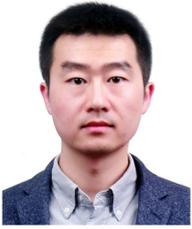
**Chao Lu** (Member, IEEE) received the B.S. degree in transport engineering from the Beijing Institute of Technology (BIT), Beijing, China, in 2009, and the Ph.D. degree in transport studies from the University of Leeds, Leeds, U.K., in 2015. In 2017, he was a Visiting Researcher with the Advanced Vehicle Engineering Centre, Cranfield University, Cranfield, U.K. He is currently an Associate Professor with the School of Mechanical Engineering, BIT. His research interests include intelligent transportation and vehicular systems, driver behavior modeling, reinforcement learning, and transfer learning and its applications.

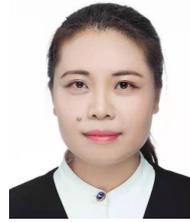
**Xuemei Chen** obtained B.E. in automobile operation engineering from the Shandong University of Technology, Zibo, in 2000 and M.S. degree from the Beijing University of Technology, Beijing, in 2003 and the Ph.D. degree in Beijing Institute of Technology, Beijing, in 2006. She is currently an Associate Professor in the Mechanical Engineering Department, the Beijing Institute of Technology and the Executive Vice president of Advanced Technology Research Institute, Beijing Institute of Technology since 2021. Her research interests include driver behavior model, autonomous vehicle decision-making and machine learning.

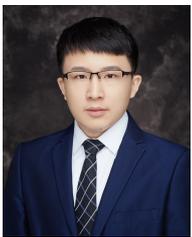
**Zirui Li** (Student Member, IEEE) received the B.S. degree from the Beijing Institute of Technology (BIT), Beijing, China, in 2019, where he is currently pursuing the Ph.D. degree in mechanical engineering. From June 2021 to July 2022, he was a Visiting Researcher with the Delft University of Technology (TU Delft). Since August 2022, he has been a Visiting Researcher with the Chair of Traffic Process Automation with the Faculty of Transportation and Traffic Sciences "Friedrich List," TU Dresden. His research interests include interactive behavior modeling, risk assessment, and the motion planning of automated vehicles.

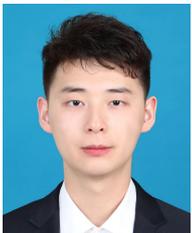
**Zhe Liu** received the B.S. degree from the Beijing Institute of Technology, Beijing, China, in 2022. He is currently pursuing the M.S. degree in mechanical engineering with the Beijing Institute of Technology, Beijing, China. His research interests include intelligent vehicles, optimal method, vehicle system dynamics, dynamic parameter estimation, motion planning and control.

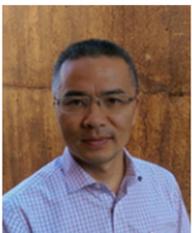
 **Jianwei Gong** (Member, IEEE) received the B.S. degree from the National University of Defense Technology, Changsha, China, in 1992, and the Ph.D. degree from the Beijing Institute of Technology (BIT), Beijing, China, in 2002. Between 2011 and 2012, he was a Visiting Scientist with the Robotic Mobility Group, Massachusetts Institute of Technology, Cambridge, MA, USA. He is currently a Professor with the School of Mechanical Engineering, BIT. His research interests include intelligent vehicle environment perception and understanding, decision making, path/motion planning, and control.